\documentclass[letterpaper]{article} %
\usepackage{aaai25}  %
\usepackage{times}  %
\usepackage{helvet}  %
\usepackage{courier}  %
\usepackage[hyphens]{url}  %
\usepackage{graphicx} %
\usepackage{natbib}  %
\usepackage{caption} %
\usepackage{algorithm}
\usepackage{algorithmic}
\usepackage{newfloat}
\usepackage{listings}
\DeclareCaptionStyle{ruled}{labelfont=normalfont,labelsep=colon,strut=off} %
\usepackage{xcolor}
\floatstyle{ruled}
\newfloat{listing}{tb}{lst}{}
\floatname{listing}{Listing}
\usepackage{amsmath}
\usepackage{booktabs}
\usepackage{xspace}
\usepackage{makecell}
\usepackage{colortbl}
\newcommand{\name}{\textsc{DomainEval}\xspace}

\newcommand{\eg}{\hbox{\emph{e.g.}}}
\newcommand{\ie}{\hbox{\emph{i.e.}}}

\title{\name: An Auto-Constructed Benchmark for\\Multi-Domain Code Generation}
\author {
    Qiming Zhu\textsuperscript{\rm 1,\rm 2}\equalcontrib,
    Jialun Cao\textsuperscript{\rm 3}\equalcontrib,
    Yaojie Lu\textsuperscript{\rm 1},
    Hongyu Lin\textsuperscript{\rm 1},\\
    Xianpei Han\textsuperscript{\rm 1},
    Le Sun\textsuperscript{\rm 1},
    Shing-Chi Cheung\textsuperscript{\rm 3}
}
\affiliations {
    \textsuperscript{\rm 1}Chinese Information Processing Laboratory, Institute of Software, Chinese Academy of Sciences, Beijing, China\\
    \textsuperscript{\rm 2}University of Chinese Academy of Sciences, Beijing, China \\
    \textsuperscript{\rm 3}The Hong Kong University of Science and Technology, Hong Kong, China \\
    \{zhuqiming2022, luyaojie, hongyu, xianpei, sunle\}@iscas.ac.cn\\
    \{jcaoap, scc\}@cse.ust.hk
}

\usepackage{bibentry}

\usepackage{appendix}
\begin{document}
\maketitle
\begin{abstract}

Code benchmarks such as HumanEval are widely adopted to evaluate capabilities of Large Language Models (LLMs), providing insights into their strengths and weaknesses. However, current benchmarks primarily exercise LLMs' capability on \textit{common coding tasks} (\eg, bubble sort, greatest common divisor), leaving \textbf{\textit{domain-specific coding tasks}} (\eg, computation, system, cryptography) unexplored. 
To fill this gap, we propose a multi-domain code benchmark, \name, designed to evaluate LLMs' coding capabilities thoroughly. 
Our pipeline works in a fully automated manner, enabling a push-bottom construction from code repositories into formatted subjects under study. 
Interesting findings are observed by evaluating 12 representative LLMs against \name. We notice that LLMs are generally good at \textbf{\textit{computation}} tasks while falling short on \textbf{\textit{cryptography} and \textit{system}} coding tasks. The performance gap can be as much as 68.94\% (80.94\% - 12.0\%) in some LLMs. We also observe that generating more samples can increase the overall performance of LLMs, while the domain bias may even increase.
The contributions of this study include a code generation benchmark dataset \name, encompassing six popular domains, a fully automated pipeline for constructing code benchmarks, and an identification of the limitations of LLMs in code generation tasks based on their performance on \name, providing directions for future research improvements.
The leaderboard is available at \url{https://domaineval.github.io/}.

\end{abstract}

\section{Introduction}

\begin{figure}[ht!]
\centering
\includegraphics[width=\linewidth]{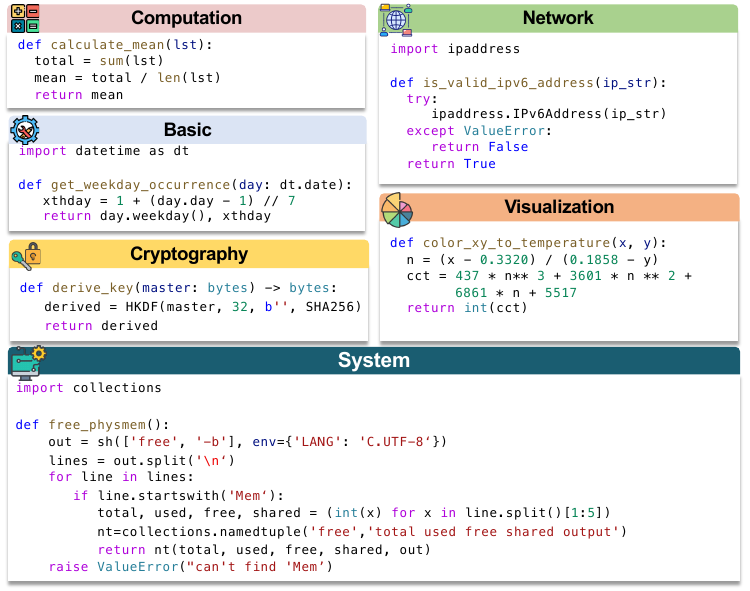}
\caption{Domain-specific Code from \name}
\label{fig:example}
\end{figure}

\begin{figure*}[ht!]
\centering
\includegraphics[width=1.0\textwidth]{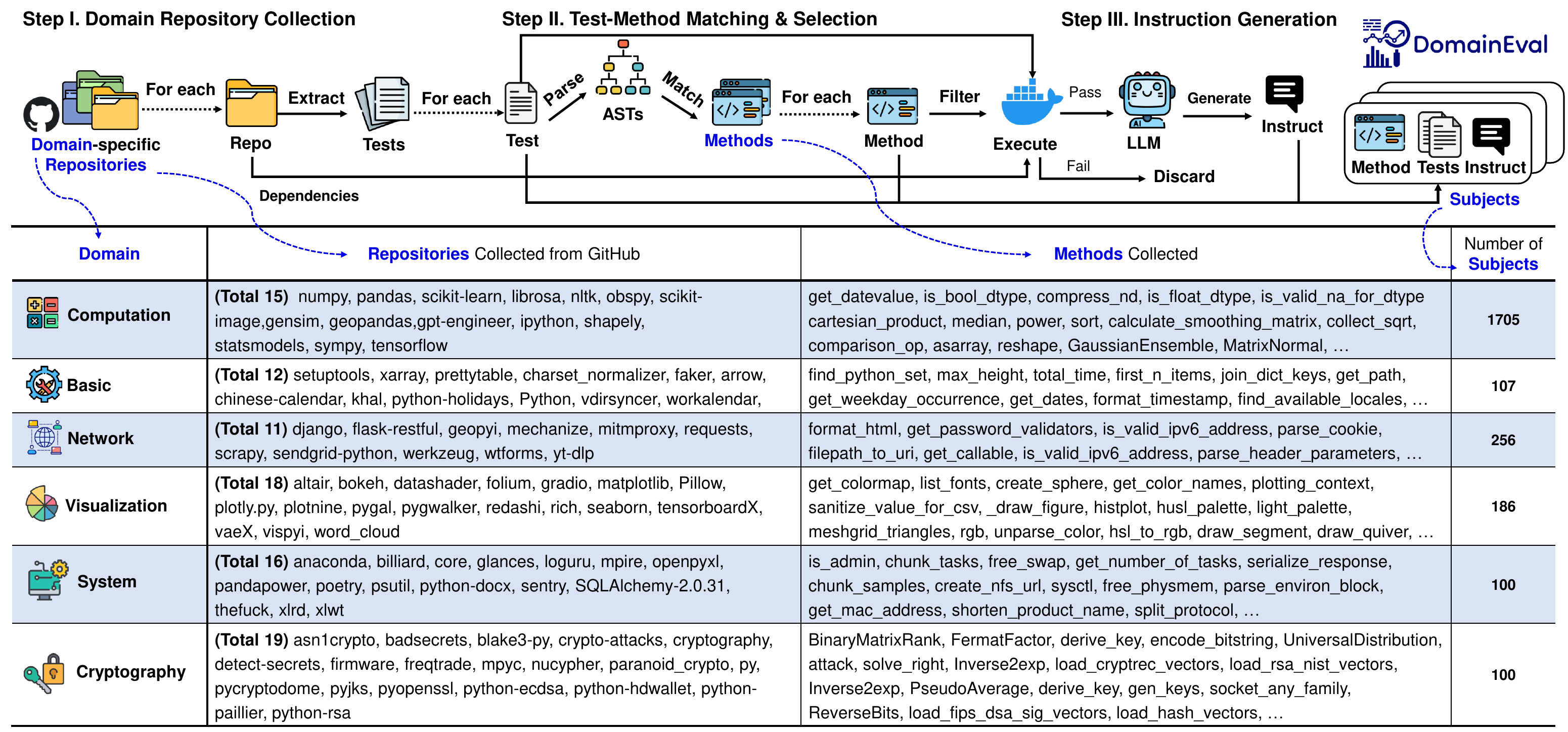} %
\caption{Pipeline of DomainEval Construction and the Domains, Repositories and Statistics in DomainEval.}
\label{fig:workflow}
\end{figure*}

Large Language Models (LLMs) have revolutionized various areas such as question answering~\cite{qa2023}, math reasoning~\cite{mathReasoning}, and especially \textit{software development}~\cite{lozhkov2024starcoder}. Stakeholders are eager to know whether and to what extent LLMs can improve development efficiency. 

To this end, a variety of {{code generation benchmarks}} such as HumanEval~\cite{chenEvaluatingLargeLanguage2021} and MBPP~\cite{austinProgramSynthesisLarge2021} have been introduced and intensively used to evaluate LLMs' coding capability. 
They primarily consist of \textbf{\textit{common coding tasks}} such as sorting an array or computing the greatest common divisor.
Furthermore, to meet more \textit{realistic} needs, more benchmarks are emerging, expanding mainly along with two perspectives. First, \textbf{\textit{linguistic diversity}}. This line of work~\cite{wang2024exploringmultilingualbiaslarge,multiPL,chai2024mceval} mitigates linguistic bias in both natural languages (\eg, English, Chinese) and programming languages (\eg, Python, Java, C/C++). Second, \textbf{\textit{code scale diversity}}. This line of work aims to scale up the code granularity from \textit{function}-level~\cite{chenEvaluatingLargeLanguage2021,yuCoderEvalBenchmarkPragmatic2024,austinProgramSynthesisLarge2021} to \textit{class}-level~\cite{du2023classeval} and \textit{repo}-level~\cite{zhangRepoCoderRepositoryLevelCode2023,javabench,li2024deveval}. 

However, if LLMs are to be applied in real-world industrial scenarios {where different product lines prioritize different domains of code}, it becomes essential to further understand these LLMs' coding capabilities across various domains. 
In other words, while existing benchmarks make significant efforts in varying natural/programming languages and code granularity, \textbf{\textit{the LLMs' ability to generate domain-specific code remains under-explored}}. 
    
To fill this gap, we introduce \name, a multi-domain code generation benchmark. It consists of 2454 subjects (\ie, reference code, description, and context), along with 5892 test cases, covering six domains of code, \ie, computation, network, basic operation, system, visualization, and cryptography. 
Figure~\ref{fig:example} shows representative examples from each domain. We can see clear functional distinctions among codes from different domains. For example, code in \textit{\textbf{computation}} involves computational tasks such as mean calculation. Code in \textit{\textbf{network}} handles network requests/communications and remote connections. Code in \textit{\textbf{cryptography}} incorporates cryptographic algorithms, encrypting plain-text with a key or conducting key recovery attacks.
    
It is noteworthy that in order to facilitate the benchmark construction and welcome future contributions to \name, we provide \textbf{\textit{a fully automated test-guided construction pipeline}}. The pipeline works in a push-bottom manner, \ie, given a code repository, it outputs a set of formatted subjects (\ie, reference code, tests, context, and description). Empowered by this pipeline, \name is with exceptional \textbf{\textit{scalability}}, capable of incorporating the ever-evolving code corpus into it. Moreover, the continuous influx of updated code through this pipeline fortifies \name against the data contamination threat~\cite{cao2024concerneddatacontamination}, thereby maintaining its integrity and novelty.

Our extensive experiments on 10+ LLMs indicate that LLMs are generally good at \textit{computation}, with an average of 82.44\% Pass@1, while falling short on \textit{cryptography} and \textit{system} domains, with an average of 33.08\% Pass@1 and 37.50\% Pass@1, respectively. The performance gap among domains can be as much as 68\%+ from \textit{Llama-2-13b-chat} model, which is observed to have the largest performance variance compared with other LLMs. We also observe that generating more samples can increase the overall performance of LLMs, while the domain bias may even increase.

The contributions can be summarized as follows. First, we introduce \name, a multi-domain code generation benchmark that consists of 2k+ subjects (\ie, description, reference code and tests) covering six domains. Second, we provide {{a fully automated test-guided construction pipeline}} to facilitate the benchmark construction and welcome future contributions. Third, we evaluate 10+ LLMs against \name and yield interesting findings.
\section{Benchmark Construction}

\begin{figure*}[tb]
    \centering
    \includegraphics[width=1.0\textwidth]{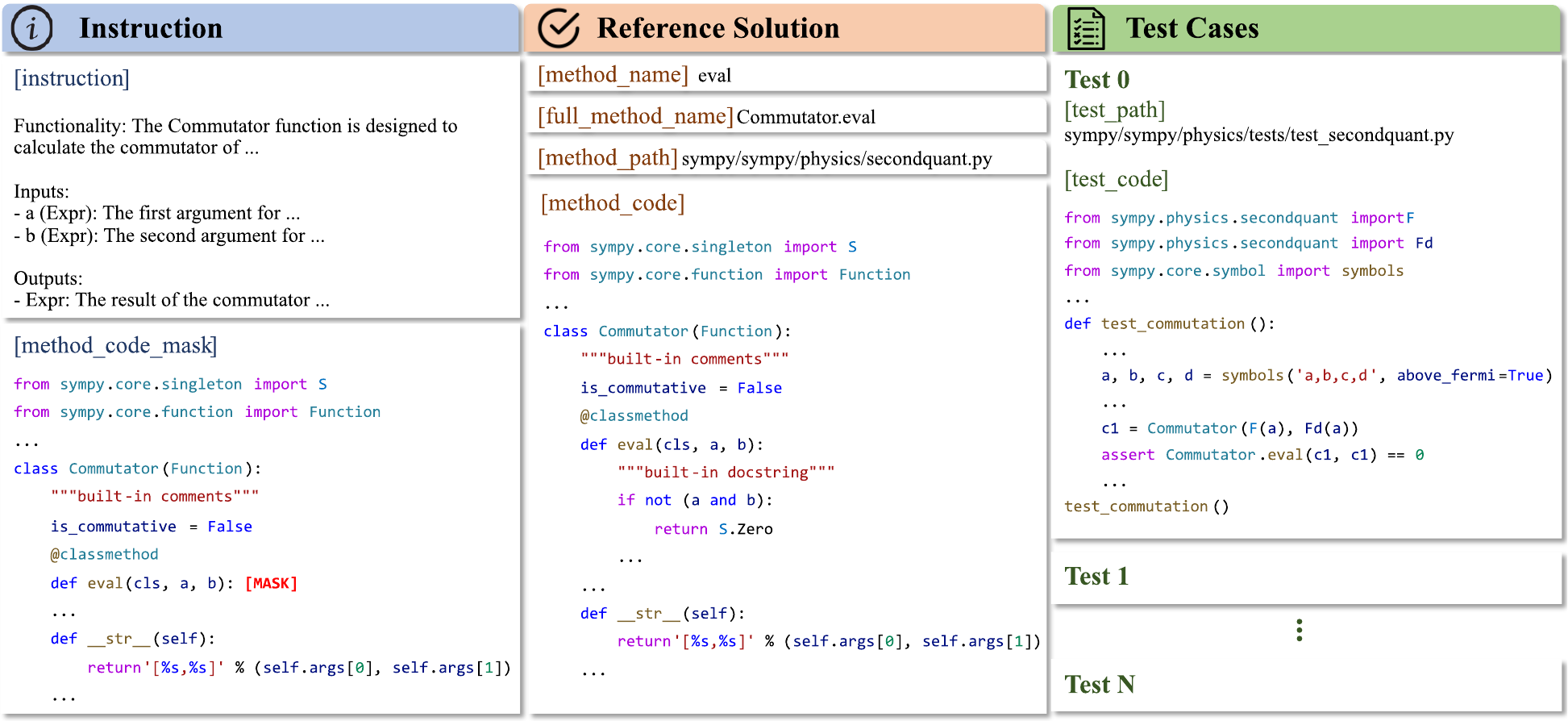} %
    \caption{A example subject in \name. Each subject consists of three components: instruction for LLM evaluation, reference solution, and a series of tests.}
    \label{fig:data display}
\end{figure*}

In this section, we introduce the whole construction pipeline of \name, as shown in Figure~\ref{fig:workflow}.
We provide a fully automated, test-guided construction algorithm that transforms code repositories into a collection of formatted subjects for LLM evaluation. 
Each subject consists of three components: instruction for LLM evaluation, reference solution, and a series of test cases, as illustrated in Figure~\ref{fig:data display}. 

The pipeline begins with the collection of raw code snippets from specific domains, which are then systematically transformed into suitable benchmark data. This process involves three key steps:
(1) \textbf{\textit{Domain Repository Collection}}: 
The first step involves collecting raw code data ($\mathbf{R}$) from selected domains.
These snippets are categorized into two groups: $\mathbf{C_0}$, containing function code, and $\mathbf{T_0}$, including tests designed to validate the correctness of the code in $\mathbf{C_0}$.
(2) \textbf{\textit{Test-Method Matching \& Selection}}:
This step aims to create a set of candidate subjects $\mathbf{S}'={(c, \mathbf{\hat{T}})}$ from the collected domain code, where each pair consists of a function code $c$ and its corresponding test cases $\mathbf{\hat{T}}$.
To achieve this, we first match each function code $c$ with its corresponding set of test cases $\mathbf{\hat{T}}$, ensuring they are compatible with the execution environments $\boldsymbol{E}$.
We then filter $\mathbf{C_0}$ to retain only those code snippets that are executable and accompanied by valid test functions, resulting in a refined set $\mathbf{C_1}$ and its associated test cases, which together form the candidate subject set $\mathbf{S}'$.
(3) \textbf{\textit{Instruction Generation}}: 
For each code snippet $c$ in $\mathbf{C_1}$, we employ a LLM to generate corresponding instruction $i=\textrm{LLM}(c)$. The instruction $i$, along with the associated function code $c$ and test cases $\mathbf{\hat{T}}$, are combined to create the final benchmark dataset, with each entry represented as $(i, c, \mathbf{\hat{T}})$, which constitutes a complete benchmark subject.

\subsection{Domain Repository Collection}

Referring to the work \citep{zhuoBigCodeBenchBenchmarkingCode2024}, we select six domains for code generation: computation, network, basic operation, system, visualization, and cryptography, to serve as the domain divisions for \name.

To ensure that \name closely aligns with the real-world code requirements of human engineers, we chose GitHub repositories as the source of raw data.
We cloned a selection of representative code repositories from GitHub, particularly those with at least 100 stars, as these are considered high-quality code data and reflect the actual needs of engineers. The code repositories used in each domain are shown in Figure~\ref{fig:workflow}.
Given that each GitHub repository represents a specific real-world application scenario, we classify the code snippets based on the repository’s domain. In practice, we use the repository’s topic labels and README files to accurately assign the code to the appropriate domain.

\subsection{Test-Method Matching \& Selection}

This section describes how to obtain reference solutions and their corresponding tests from code repositories to construct candidate subjects for LLM evaluation and the automated construction pipeline.

Given that code snippets from GitHub repositories are written by human engineers in real-world production environments, they tend to be more complex, with more dependencies and higher levels of encapsulation than standalone functions in some code datasets.
This complexity makes it challenging to select function code that can serve as benchmark data while simultaneously acquiring corresponding test cases to form complete candidate subjects.

To address this challenge, we propose a test-method tracing strategy aimed at constructing candidate subjects for code benchmarking. 
Our approach has two main steps:
First, we search the code repository for tests related to the reference code and perform Test-method Matching, where each function code $c$ is automatically paired with its corresponding tests $\mathbf{\hat{T}}$.
Second, to ensure the final candidates are suitable for LLM evaluation and can smoothly pass through the automated construction pipeline, we filter the matched results using three criteria: executable, significant, and appropriate difficulty.

\subsubsection{Test-Method Matching}
To match the function code with corresponding tests and package them into candidate subjects, we start with the test code and trace it back to the associated function code. 
Specifically, we search for Python code files in repositories and use the library \textit{ast} to parse Python code into an abstract syntax tree.
Then, we sequentially traverse the nodes of the syntax tree, extracting all functions and their class context (if present). 
Next, we select test code snippets based on two heuristic rules: First, the function or class name of a test should contain \texttt{test} or \texttt{Test}. Second, the test code snippet should contain an \texttt{assert} statement.

To match the selected test code with the corresponding function code, we identify \textit{ast.Call} nodes as function calls.
To retrieve the implementation of the called function, we consider two scenarios:
If the function and test are in the same file, we traverse the abstract syntax tree to locate and unparse the function node.
If the function call spans files, we use recursive path matching by analyzing Python’s import behaviors until the specific file is found.
Specifically, we continuously retrieve and replace the name of the next level path based on the content of \texttt{\_\_init\_\_.py} and the import statement in \texttt{from... import... as...} format, until the specific Python file is located.
Once located, the process is the same as in the first scenario, using \textit{ast} to parse and identify the function node.

In real-world code repositories, the correlation between tests and function code is often not one-to-one.
To address this, we identify all functions within the test code and pair them with the corresponding function code.
The function code is then used as the reference solution, and we group all related tests into a test suite. 
This process allows us to package function code snippets from GitHub repositories with their corresponding test cases, creating candidate subjects.

\subsubsection{Test-Method Selection}

After packaging the candidate subjects, we continue to construct our benchmark dataset. We notice that not every function can directly convert to benchmark data suitable for LLM evaluation. 
Therefore, to facilitate the automatic construction of code benchmarks, we impose three criteria on the candidate subject.

First, \textbf{\textit{Executable}}. To use Pass@k~\citep{chenEvaluatingLargeLanguage2021} for evaluation, reference solution code must be executable to verify semantic consistency with generated code against test cases.
To ensure the security of our benchmark data, we utilize a sandbox to isolate code execution and maintain a list of banned keywords~\cite{xie2024codebenchgen}, as detailed in the Appendix.
If any of these prohibited keywords are detected, we consider the execution and evaluation of such code to be potentially risky and consequently discard it.
The test environment first consists of a basic Python environment and necessary packages.
To execute a piece of code, it needs adequate dependencies, so we concatenate the required context (e.g., import statements, class context, static variables) for the function. 
After preparing the context, we run the function code along with its tests.

Second, \textbf{\textit{Significant}}. The code used for LLM evaluation should be important and meaningful, playing a critical role in real-world production scenarios.
For example, \texttt{\_\_init\_\_} functions primarily involve repetitive variable assignments, which are mechanical and lack significance, failing to reflect the capabilities of LLMs in code generation. 
In contrast, human engineers tend to write tests for code that implements critical functionality. Therefore, we extract the code with tests as candidates. 

Third, \textbf{\textit{Appropriate Difficulty}}. The number of lines of reference code is one of the direct indicators of the complexity of the code generation task. We set a limit on the function implementation of reference code, \ie, the standard answer, used in the task. We restrict them to between 3 and 100 lines. On the one hand, functions with fewer than three lines typically have overly simple logic \citep{yuCoderEvalBenchmarkPragmatic2024} and do not effectively reveal the shortcomings of LLMs. On the other hand, functions exceeding 100 lines may contain overly complex logic, which can present significant challenges for evaluation.
These challenges include exceeding LLMs' context limitations or overwhelming their information capacity, making it difficult to generate precise instructions in subsequent steps and thus hindering the automatic construction pipeline.

\subsection{Instruction Generation}

\begin{figure}[tb]
\centering
\includegraphics[width=0.94\columnwidth]{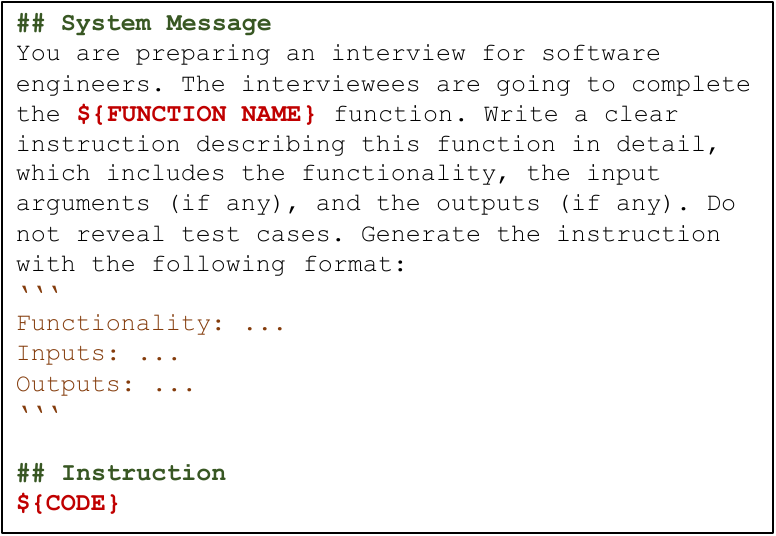} %
\caption{Instruction Generation prompt for LLM}
\label{fig:Instruction Generation prompt for LLM}
\end{figure}

After filtering the candidate subjects, our subjects still lack the \textit{instruction} field. To ensure the reproducibility of the dataset construction process, we employ open-source LLM, \ie \textit{Qwen2-72B-Instruct-GPTQ-Int4}, for data generation. We utilize a \textbf{\textit{dialogue prompt template}}, as depicted in Figure~\ref{fig:Instruction Generation prompt for LLM}, to guide the LLM in generating detailed \textit{Instruction} fields for each validated subject. These \textit{instruction} fields comprehensively describe the function, including its purpose, input arguments, and expected outputs, which serve as the input for evaluating the code generation capabilities of LLMs. By leveraging LLMs, we can generate natural language descriptions that outline the desired functionality, inputs, and outputs of the code \cite{xie2024codebenchgen}.

Finally, to construct the template for code generation, we mask the function code, following prior work~\cite{xie2024codebenchgen}, by \texttt{[MASK]} as shown in the \textit{method\_code\_mask} field in Figure~\ref{fig:data display}, and instruct LLMs to complete the masked segments during the evaluation.

\subsection{Benchmark Statistics}

The above pipeline generates instructions, reference code, and a series of tests shown in Figure~\ref{fig:data display}, which combine to form complete subjects. 
As a result, \name consists of 2454 code subjects.
Figure~\ref{fig:workflow} shows the number of subjects constructed from repositories across different domains. Overall, the number of subjects from each domain is at least 100. The \textit{computation} domain encompasses the greatest number of subjects, with 1705 subjects, compared to 100 $\sim$ 256 subjects in other domains. This is because the computation-related repositories offer more function code and also include a larger number of test cases to ensure the accuracy of each computational operation.

\begin{figure}[tb]
\centering
\includegraphics[width=0.9\columnwidth]{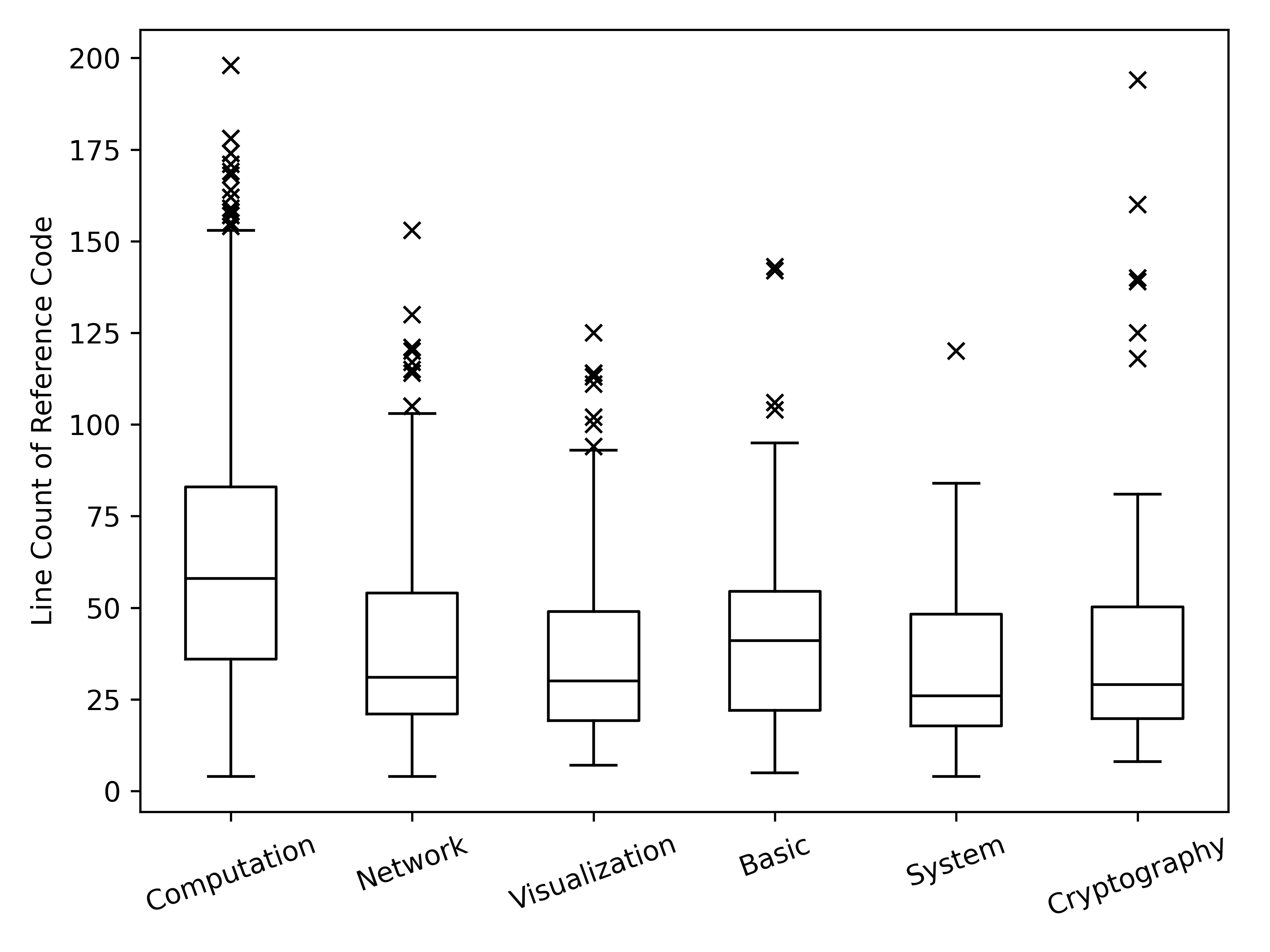} %
\caption{Distribution of line counts for the reference code (context included) across domains in \name.}
\label{fig:line counts}
\end{figure}

Figure~\ref{fig:line counts} illustrates the distribution of lines of reference code within \name. Overall, the lines of reference code (context included) across six domains are similar, ranging from 4 to 198, with an average of 55.69. In particular, code in \textit{computation} has slightly more lines of code than that in other domains, with an average of 63.20 lines of code compared with 33.95 $\sim$ 42.03 in other domains.

\section{Experiments}

\subsection{Experiment Setup}

\noindent \textbf{Studied LLMs.}
We assess 12 representative instruction-tuned LLMs against \name, including GPT-3.5-turbo, GPT-4o-mini~\cite{brown2020language,achiam2023gpt}, Qwen2~\cite{yang2024qwen2}, Phi-3~\cite{abdin2024phi}), DeepSeek-Coder series~\cite{zhu2024deepseek,guo2024deepseek}, Llama2~\cite{touvron2023llama} and CodeLlama series~\cite{roziere2023code}, CodeQwen1.5~\cite{bai2023qwen} with size of open-source models from 6.7B to 72B. These models exhibit proficiency in following instructions and delivering appropriately formatted responses.

\begin{figure}[tb]
\centering
\includegraphics[width=0.95\columnwidth]{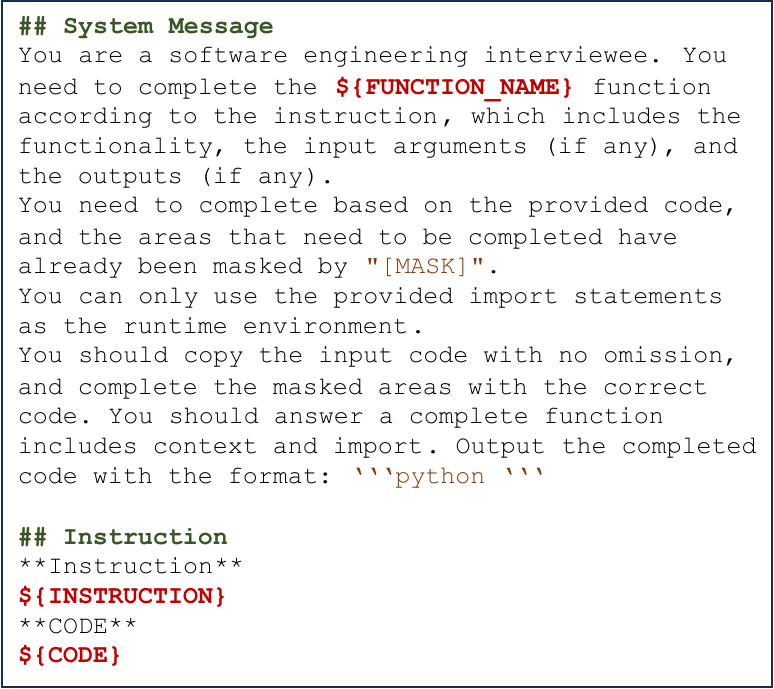} %
\caption{Code Generation Task Prompt}
\label{fig:Code Generation Task Prompt}
\end{figure}

\noindent \textbf{Evaluation Metrics.}
Our evaluation uses the unbiased version of Pass@k \citep{chenEvaluatingLargeLanguage2021} to accurately assess the functional correctness of code snippets generated by LLMs. Following prior work \citep{zhuoBigCodeBenchBenchmarkingCode2024}, we report Pass@1 and Pass@5 for the experiment in zero-shot setting and use macro-average as scores.
For Pass@1 metric, we use greedy decoding, \ie set temperature to $0.0$.
For Pass@5 metric, we opt for the minimum sample size $N=5$ and maintain temperature at $0.2$ and top-p at $0.95$. For code generation tasks, we use \textit{torch.bfloat16} when loading LLMs and the same prompt detailed in Figure~\ref{fig:Code Generation Task Prompt}. 

\noindent \textbf{Evaluation Process.} During the evaluation, to prevent LLMs from failing execution due to omitted import statements, which is a tolerable flaw but could potentially distort assessment results, we implement a corrective measure, \ie, completing the missing dependencies based on the import scenario mentioned in the instruction.

\begin{table*}[h!]
\centering
\small
    \begin{tabular}{l|c|cccccc|cc}
    \toprule
    \textbf{Pass@1} (Greedy Search N=1) & \textbf{Size} & \textbf{Comp} & \textbf{Network} & \textbf{Visual} & \textbf{Basic} & \textbf{System} & \textbf{Crypt} & \textbf{Mean} & \textbf{Std} \\
    \midrule
    GPT-4o-mini & \textbackslash{} & \cellcolor[rgb]{ .569,  .667,  .875} 90.38  & \cellcolor[rgb]{ .682,  .753,  .91} 70.31  & \cellcolor[rgb]{ .741,  .8,  .925} 59.68  & \cellcolor[rgb]{ .686,  .761,  .91} 69.16  & \cellcolor[rgb]{ .788,  .835,  .941} 51.00  & \cellcolor[rgb]{ .831,  .871,  .953} 43.00  & \cellcolor[rgb]{ .957,  .702,  .51} 63.92  & \cellcolor[rgb]{ .949,  .976,  .929} 16.68  \\
    GPT-3.5-turbo & \textbackslash{} & \cellcolor[rgb]{ .608,  .698,  .886} 83.40  & \cellcolor[rgb]{ .745,  .804,  .925} 58.98  & \cellcolor[rgb]{ .8,  .843,  .941} 48.92  & \cellcolor[rgb]{ .761,  .816,  .933} 56.07  & \cellcolor[rgb]{ .89,  .918,  .969} 32.00  & \cellcolor[rgb]{ .898,  .922,  .973} 31.00  & \cellcolor[rgb]{ .98,  .863,  .769} 51.73  & \cellcolor[rgb]{ .859,  .933,  .804} 19.50  \\
    Qwen2-72B-Instruct-GPTQ-Int4 & 72B   & \cellcolor[rgb]{ .588,  .682,  .882} 86.86  & \cellcolor[rgb]{ .702,  .769,  .914} 66.80  & \cellcolor[rgb]{ .796,  .843,  .941} 49.46  & \cellcolor[rgb]{ .686,  .761,  .91} 69.16  & \cellcolor[rgb]{ .843,  .878,  .957} 41.00  & \cellcolor[rgb]{ .871,  .898,  .965} 36.00  & \cellcolor[rgb]{ .976,  .796,  .667} 58.21  & \cellcolor[rgb]{ .863,  .937,  .808} 19.39  \\
    DeepSeek-Coder-33b-instruct & 33B   & \cellcolor[rgb]{ .608,  .694,  .886} 83.93  & \cellcolor[rgb]{ .714,  .78,  .918} 64.45  & \cellcolor[rgb]{ .788,  .839,  .941} 50.54  & \cellcolor[rgb]{ .737,  .8,  .925} 59.81  & \cellcolor[rgb]{ .816,  .859,  .949} 46.00  & \cellcolor[rgb]{ .875,  .906,  .965} 35.00  & \cellcolor[rgb]{ .98,  .816,  .698} 56.62  & \cellcolor[rgb]{ .941,  .973,  .918} 16.94  \\
    DeepSeek-Coder-V2-Lite-Instruct & 16B   & \cellcolor[rgb]{ .596,  .686,  .882} 86.04  & \cellcolor[rgb]{ .725,  .788,  .922} 62.11  & \cellcolor[rgb]{ .792,  .839,  .941} 50.00  & \cellcolor[rgb]{ .71,  .776,  .918} 65.42  & \cellcolor[rgb]{ .843,  .878,  .957} 41.00  & \cellcolor[rgb]{ .859,  .89,  .961} 38.00  & \cellcolor[rgb]{ .98,  .812,  .69} 57.10  & \cellcolor[rgb]{ .91,  .957,  .875} 17.92  \\
    DeepSeek-Coder-6.7b-instruct & 6.7B  & \cellcolor[rgb]{ .608,  .698,  .886} 83.52  & \cellcolor[rgb]{ .745,  .804,  .925} 58.98  & \cellcolor[rgb]{ .816,  .859,  .949} 45.70  & \cellcolor[rgb]{ .749,  .808,  .929} 57.94  & \cellcolor[rgb]{ .871,  .898,  .965} 36.00  & \cellcolor[rgb]{ .847,  .882,  .957} 40.00  & \cellcolor[rgb]{ .984,  .851,  .757} 53.69  & \cellcolor[rgb]{ .929,  .969,  .902} 17.32  \\
    CodeLlama-34b-Instruct & 34B   & \cellcolor[rgb]{ .651,  .729,  .898} 76.07  & \cellcolor[rgb]{ .737,  .796,  .925} 60.16  & \cellcolor[rgb]{ .835,  .875,  .953} 41.94  & \cellcolor[rgb]{ .765,  .82,  .933} 55.14  & \cellcolor[rgb]{ .875,  .906,  .965} 35.00  & \cellcolor[rgb]{ .898,  .922,  .973} 31.00  & \cellcolor[rgb]{ .988,  .894,  .827} 49.89  & \cellcolor[rgb]{ .937,  .973,  .914} 17.09  \\
    CodeLlama-13b-Instruct & 13B   & \cellcolor[rgb]{ .627,  .71,  .894} 80.29  & \cellcolor[rgb]{ .725,  .788,  .922} 62.11  & \cellcolor[rgb]{ .835,  .871,  .953} 42.47  & \cellcolor[rgb]{ .745,  .804,  .925} 58.88  & \cellcolor[rgb]{ .882,  .91,  .969} 34.00  & \cellcolor[rgb]{ .918,  .937,  .976} 27.00  & \cellcolor[rgb]{ .988,  .886,  .812} 50.79  & \cellcolor[rgb]{ .847,  .929,  .784} 19.90  \\
    CodeLlama-7b-Instruct & 7B    & \cellcolor[rgb]{ .643,  .725,  .898} 77.13  & \cellcolor[rgb]{ .733,  .796,  .925} 60.55  & \cellcolor[rgb]{ .827,  .867,  .953} 43.55  & \cellcolor[rgb]{ .78,  .831,  .937} 52.34  & \cellcolor[rgb]{ .871,  .898,  .965} 36.00  & \cellcolor[rgb]{ .89,  .918,  .969} 32.00  & \cellcolor[rgb]{ .988,  .89,  .82} 50.26  & \cellcolor[rgb]{ .945,  .976,  .925} 16.82  \\
    CodeQwen1.5-7B-Chat & 7B    & \cellcolor[rgb]{ .6,  .69,  .886} 85.16  & \cellcolor[rgb]{ .733,  .792,  .925} 60.94  & \cellcolor[rgb]{ .804,  .851,  .945} 47.85  & \cellcolor[rgb]{ .733,  .796,  .925} 60.75  & \cellcolor[rgb]{ .863,  .894,  .961} 37.00  & \cellcolor[rgb]{ .863,  .894,  .961} 37.00  & \cellcolor[rgb]{ .984,  .839,  .733} 54.78  & \cellcolor[rgb]{ .898,  .953,  .855} 18.31  \\
    Phi-3-medium-4k-instruct & 14B   & \cellcolor[rgb]{ .651,  .733,  .902} 75.54  & \cellcolor[rgb]{ .737,  .796,  .925} 60.16  & \cellcolor[rgb]{ .82,  .863,  .949} 45.16  & \cellcolor[rgb]{ .729,  .792,  .922} 61.68  & \cellcolor[rgb]{ .835,  .875,  .953} 42.00  & \cellcolor[rgb]{ .875,  .906,  .965} 35.00  & \cellcolor[rgb]{ .984,  .855,  .765} 53.26  & 15.10  \\
    Llama-2-13b-chat & 13B   & \cellcolor[rgb]{ .624,  .71,  .89} 80.94  & \cellcolor[rgb]{ .776,  .827,  .937} 53.12  & \cellcolor[rgb]{ .875,  .906,  .965} 34.95  & \cellcolor[rgb]{ .82,  .863,  .949} 44.86  & \cellcolor[rgb]{ .965,  .973,  .992} 19.00  & 12.00  & 40.81  & \cellcolor[rgb]{ .678,  .847,  .553} 24.97  \\
    \midrule
    \textbf{Average} & \textbackslash{} & \cellcolor[rgb]{ .616,  .702,  .89} 82.44  & \cellcolor[rgb]{ .729,  .792,  .922} 61.56  & \cellcolor[rgb]{ .812,  .855,  .945} 46.69  & \cellcolor[rgb]{ .741,  .8,  .925} 59.27  & \cellcolor[rgb]{ .863,  .894,  .961} 37.50  & \cellcolor[rgb]{ .886,  .914,  .969} 33.08  & 53.42  & 18.33  \\
    \midrule
    \addlinespace
    \midrule
    \textbf{Pass@5} (Sampling Search N=5) & \textbf{Size} & \textbf{Comp} & \textbf{Network} & \textbf{Visual} & \textbf{Basic} & \textbf{System} & \textbf{Crypt} & \textbf{Mean} & \textbf{Std} \\
    \midrule
    GPT-4o-mini & \textbackslash{} & \cellcolor[rgb]{ .569,  .667,  .875} 91.26  & \cellcolor[rgb]{ .686,  .757,  .91} 72.66  & \cellcolor[rgb]{ .753,  .808,  .929} 61.83  & \cellcolor[rgb]{ .694,  .765,  .914} 71.03  & \cellcolor[rgb]{ .78,  .831,  .937} 57.00  & \cellcolor[rgb]{ .831,  .871,  .953} 49.00  & \cellcolor[rgb]{ .957,  .702,  .51} 67.13  & 14.75  \\
    GPT-3.5-turbo & \textbackslash{} & \cellcolor[rgb]{ .596,  .686,  .882} 87.33  & \cellcolor[rgb]{ .745,  .804,  .925} 62.89  & \cellcolor[rgb]{ .812,  .855,  .945} 52.15  & \cellcolor[rgb]{ .757,  .812,  .929} 60.75  & \cellcolor[rgb]{ .91,  .929,  .976} 36.00  & \cellcolor[rgb]{ .922,  .941,  .98} 34.00  & \cellcolor[rgb]{ .984,  .871,  .788} 55.52  & \cellcolor[rgb]{ .831,  .922,  .765} 19.74  \\
    Qwen2-72B-Instruct-GPTQ-Int4 & 72B   & \cellcolor[rgb]{ .576,  .675,  .878} 90.15  & \cellcolor[rgb]{ .698,  .765,  .914} 70.70  & \cellcolor[rgb]{ .796,  .843,  .941} 54.84  & \cellcolor[rgb]{ .678,  .753,  .906} 73.83  & \cellcolor[rgb]{ .824,  .863,  .949} 50.00  & \cellcolor[rgb]{ .847,  .882,  .957} 46.00  & \cellcolor[rgb]{ .965,  .745,  .58} 64.25  & \cellcolor[rgb]{ .929,  .969,  .898} 16.90  \\
    DeepSeek-Coder-33b-instruct & 33B   & \cellcolor[rgb]{ .58,  .675,  .878} 89.79  & \cellcolor[rgb]{ .698,  .765,  .914} 70.70  & \cellcolor[rgb]{ .792,  .839,  .941} 55.38  & \cellcolor[rgb]{ .714,  .776,  .918} 68.22  & \cellcolor[rgb]{ .78,  .831,  .937} 57.00  & \cellcolor[rgb]{ .875,  .902,  .965} 42.00  & \cellcolor[rgb]{ .965,  .753,  .588} 63.85  & \cellcolor[rgb]{ .949,  .976,  .925} 16.34  \\
    DeepSeek-Coder-V2-Lite-Instruct & 16B   & \cellcolor[rgb]{ .584,  .678,  .882} 88.91  & \cellcolor[rgb]{ .729,  .792,  .922} 65.62  & \cellcolor[rgb]{ .8,  .847,  .945} 53.76  & \cellcolor[rgb]{ .714,  .776,  .918} 68.22  & \cellcolor[rgb]{ .831,  .871,  .953} 49.00  & \cellcolor[rgb]{ .859,  .894,  .961} 44.00  & \cellcolor[rgb]{ .969,  .784,  .643} 61.59  & \cellcolor[rgb]{ .945,  .976,  .925} 16.35  \\
    DeepSeek-Coder-6.7b-instruct & 6.7B  & \cellcolor[rgb]{ .58,  .675,  .878} 89.79  & \cellcolor[rgb]{ .741,  .8,  .925} 63.67  & \cellcolor[rgb]{ .792,  .839,  .941} 55.38  & \cellcolor[rgb]{ .718,  .78,  .918} 67.29  & \cellcolor[rgb]{ .831,  .871,  .953} 49.00  & \cellcolor[rgb]{ .859,  .894,  .961} 44.00  & \cellcolor[rgb]{ .969,  .784,  .643} 61.52  & \cellcolor[rgb]{ .945,  .976,  .925} 16.36  \\
    CodeLlama-34b-Instruct & 34B   & \cellcolor[rgb]{ .608,  .698,  .886} 85.10  & \cellcolor[rgb]{ .741,  .8,  .925} 63.28  & \cellcolor[rgb]{ .835,  .871,  .953} 48.39  & \cellcolor[rgb]{ .745,  .804,  .929} 62.62  & \cellcolor[rgb]{ .878,  .906,  .965} 41.00  & \cellcolor[rgb]{ .875,  .902,  .965} 42.00  & \cellcolor[rgb]{ .98,  .851,  .749} 57.07  & \cellcolor[rgb]{ .929,  .969,  .902} 16.83  \\
    CodeLlama-13b-Instruct & 13B   & \cellcolor[rgb]{ .58,  .675,  .878} 89.85  & \cellcolor[rgb]{ .729,  .792,  .922} 65.62  & \cellcolor[rgb]{ .816,  .855,  .949} 51.61  & \cellcolor[rgb]{ .722,  .788,  .922} 66.36  & \cellcolor[rgb]{ .898,  .922,  .973} 38.00  & \cellcolor[rgb]{ .918,  .937,  .976} 35.00  & \cellcolor[rgb]{ .98,  .839,  .733} 57.74  & \cellcolor[rgb]{ .804,  .906,  .725} 20.55  \\
    CodeLlama-7b-Instruct & 7B    & \cellcolor[rgb]{ .596,  .69,  .886} 86.80  & \cellcolor[rgb]{ .741,  .8,  .925} 63.67  & \cellcolor[rgb]{ .816,  .855,  .949} 51.61  & \cellcolor[rgb]{ .733,  .796,  .925} 64.49  & \cellcolor[rgb]{ .867,  .898,  .961} 43.00  & \cellcolor[rgb]{ .886,  .914,  .969} 40.00  & \cellcolor[rgb]{ .976,  .831,  .722} 58.26  & \cellcolor[rgb]{ .914,  .961,  .882} 17.28  \\
    CodeQwen1.5-7B-Chat & 7B    & \cellcolor[rgb]{ .573,  .671,  .878} 91.03  & \cellcolor[rgb]{ .737,  .796,  .925} 64.06  & \cellcolor[rgb]{ .792,  .839,  .941} 55.38  & \cellcolor[rgb]{ .714,  .776,  .918} 68.22  & \cellcolor[rgb]{ .855,  .886,  .961} 45.00  & \cellcolor[rgb]{ .875,  .902,  .965} 42.00  & \cellcolor[rgb]{ .973,  .792,  .659} 60.95  & \cellcolor[rgb]{ .89,  .949,  .847} 17.95  \\
    Phi-3-medium-4k-instruct & 14B   & \cellcolor[rgb]{ .608,  .698,  .886} 85.10  & \cellcolor[rgb]{ .718,  .78,  .918} 67.58  & \cellcolor[rgb]{ .796,  .843,  .941} 54.30  & \cellcolor[rgb]{ .718,  .78,  .918} 67.29  & \cellcolor[rgb]{ .843,  .878,  .957} 47.00  & \cellcolor[rgb]{ .859,  .894,  .961} 44.00  & \cellcolor[rgb]{ .973,  .796,  .659} 60.88  & \cellcolor[rgb]{ .976,  .992,  .969} 15.45  \\
    Llama-2-13b-chat & 13B   & \cellcolor[rgb]{ .592,  .686,  .882} 87.68  & \cellcolor[rgb]{ .788,  .835,  .941} 55.86  & \cellcolor[rgb]{ .886,  .914,  .969} 39.78  & \cellcolor[rgb]{ .831,  .871,  .953} 48.60  & \cellcolor[rgb]{ .973,  .976,  .992} 26.00  & 21.00  & 46.49  & \cellcolor[rgb]{ .678,  .847,  .553} 24.10  \\
    \midrule
    \textbf{Average} & \textbackslash{} & \cellcolor[rgb]{ .588,  .682,  .882} 88.57  & \cellcolor[rgb]{ .729,  .792,  .922} 65.53  & \cellcolor[rgb]{ .808,  .851,  .945} 52.87  & \cellcolor[rgb]{ .729,  .792,  .922} 65.58  & \cellcolor[rgb]{ .855,  .89,  .961} 44.83  & \cellcolor[rgb]{ .882,  .91,  .969} 40.25  & 59.60  & 17.72  \\
    \bottomrule
    \end{tabular}%
\caption{\textbf{Pass@1} and \textbf{Pass@5} of LLMs against \name. We use \textit{Comp} for \textit{computation}, \textit{Visual} for \textit{visualization}, and \textit{Crypt} for \textit{cryptography}. \textbf{Mean} represents the macro-average of Pass@k across different domains, which is used to reflect the overall performance of LLMs. \textbf{Std} indicates the standard deviation of Pass@k across different domains, which is used to reflect the difference degree in performance of LLM across various domains. To highlight the differences, we use color scales.}
\label{tab:result_pass@k}
\end{table*}

\subsection{Overall Result}
Table~\ref{tab:result_pass@k} shows LLMs' Pass@1 and Pass@5 against \name. Columns plotted in blue show the Pass@1/5 values; the bluer, the larger. The columns in orange and green highlight the average (``Mean'') and standard deviation (``Std'') of the corresponding rows, respectively. Overall, the average performance across studied LLMs is similar, ranging from 49.89\% $\sim$ 67.13\%. At the same time, the performance across domains varies, \ie, the performance in \textit{Computation} reaches the top among all LLMs, with an average of 82.44\% Pass@1 and 88.57\% Pass@5, while the worst scores are observed in Cryptography domain, with 33.08\% Pass@1 and 40.25\% Pass@5. In other words, the performance gaps between different domains are unignorable.

\subsection{Domain Biases}
From Table~\ref{tab:result_pass@k}, we can see significant gaps across six domains. Horizontally, LLMs are generally \textbf{\textit{good at \textit{computation} tasks}} while \textbf{\textit{falling short on \textit{cryptography} and \textit{system}}} coding tasks. In particular, LLMs excel in \textit{computation} domain, where Pass@k metrics all exceed 75\%, with some reaching over 90\%. When it comes to \textit{cryptography} and \textit{system} domains, LLMs exhibit significantly lower performance, with average Pass@1 of 33.08\% and 37.50\%, respectively. The performance gap can be as much as 68.94\% (80.94\% - 12.0\%) Pass@1 in Llama-2-13b-chat. Vertically, all LLMs exhibit similar domain biases, \ie, LLMs universally show consistent performance gaps with \textbf{\textit{a shared trend of strengths and weaknesses}}.

In addition, a recent work~\citep{zhuoBigCodeBenchBenchmarkingCode2024} also explored the coding capability (\ie, using APIs to implement domain-specific code) across various domains and concluded that LLMs are good at \textit{cryptography} domain. Our conclusion does not conflict with theirs because we generate the domain-specific code while they invoke the domain-specific APIs. In other words, \textit{\textbf{being good at calling APIs in a domain does not mean being good at implementing code in the domain}}. Therefore, our finding serves as a supplement to previous work~\citep{zhuoBigCodeBenchBenchmarkingCode2024}.

\subsection{LLMs Biases}
Among 12 studied LLMs, the closed-source model\textbf{\textit{ \textit{GPT-4o-mini} exhibits the average highest performance}}, with a 67.13\% Pass@5. \textit{Qwen2-72B-Instruct-GPTQ-Int4} has the best overall performance among open-source models, with a 64.25\% Pass@5. 
Moreover, considering the variation across domains, \textbf{\textit{GPT-4o-mini} exhibits the most stable performance}, with a 14.75 standard deviation in Pass@5, compared with 15.45 $\sim$ 24.10 of other LLMs. 

Notably, \textit{CodeLlama-13b}, which fine-tuned from \textit{Llama-2-13b}, achieves an 11.25\% (57.74\% - 46.49\%) average improvement, while the deviation across domains still remains. It indicates that \textbf{\textit{although fine-tuning can bring about overall improvement, while the domain gaps still exist.}}

\subsection{Impact of Generated Samples}
Finally, we analyze the impact of generated samples (\ie, \textit{(Pass@1 Greedy Search N=1)} with sub-table \textit{(Pass@5 Sampling Search N=5)}). From Table~\ref{tab:result_pass@k}, we can see that after increasing the number of samples from 1 to 5, the average performance increases from {53.42\% to 59.60\%}, with consistent improvements on all six domains. Yet, in terms of standard deviation (the smaller, the less bias, the better), there is little improvement, from an average {18.33 to 17.72}. What is worse, \textit{CodeLlama-13b-instruct} even observes an increased deviation, from 19.90 to 20.55, indicating a more bias as generation goes on. In other words, \textbf{\textit{generating more samples can increase the overall performance, while the domain bias may even increase.}}

\subsection{Case Study}

Despite impressive performance exhibited by LLMs in \textit{computation} domain, their shortcomings in other domains cannot be overlooked. This claim is supported by two indicative cases from \textit{cryptography} and \textit{system} domains, which highlight the challenges faced by LLMs. LLMs need to acquire more background knowledge to enhance their code generation capabilities in specific vertical domains.

\begin{figure}[tb]
\centering
\includegraphics[width=1.0\columnwidth]{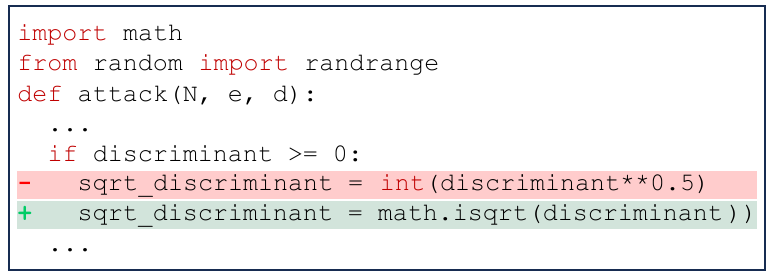} %
\caption{A flawed response from \textit{GPT-4o-mini} in \textit{cryptography} domain.}
\label{fig:case_Crypt}
\end{figure}
The response presented in Figure~\ref{fig:case_Crypt} is from \textit{GPT-4o-mini}, which is considered as the premier model within \textit{cryptography} domain. The error in its response is \textit{failed: int too large to convert to float}. Similar errors are also observed in some responses from \textit{DeepSeek} and \textit{Qwen} series models.

The context of this subject is a classical \textit{cryptography} scenario, the attack method targeting the RSA encryption algorithm. This involves the recovery of the two prime factors, $p$ and $q$, of the RSA modulus $N$, given the public key (consisting of the modulus $N$ and the public exponent $e$) and the private key (the private exponent $d$). 
When encrypting, in order to enhance the security and impede decryption attempts, the selected $p$ and $q$ by human are both extremely large prime numbers. Although the instruction mentioned keywords such as RSA, the model does not realize that in the context of \textit{cryptography}, rounding the square root of a large number cannot be directly converted to \textit{**0.5}, and instead \textit{math.isqrt} should be used to avoid \textit{OverflowError}.

\begin{figure}[tb]
\centering
\includegraphics[width=1.0\columnwidth]{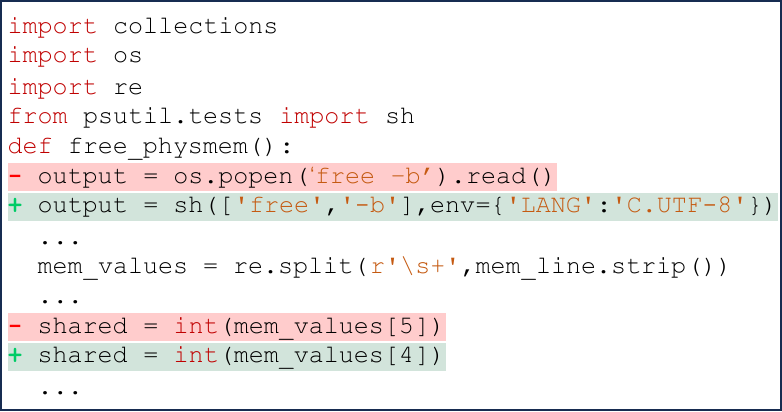} %
\caption{A flawed response from \textit{GPT-4o-mini} in \textit{system} domain.}
\label{fig:case_System}
\end{figure}
In the same way, although \textit{GPT-4o-mini} exhibits the best performance in \textit{system} domain, Figure~\ref{fig:case_System} shows a flawed response from it. The function is designed to parse the output of \textit{free} command (run with \textit{-b} option) to determine the physical memory state on a Linux system. However, the model fails to ensure that the output format of \textit{free} command remains consistent across different language and regional settings, resulting in incorrect string matching during subsequent parsing. Similar errors are observed in models such as \textit{CodeQwen1.5-7B-Chat}, \textit{Qwen2-72B-Instruct-GPTQ-Int4}, \textit{Phi-3-medium-4k-instruct}, the \textit{DeepSeek} series, and the \textit{CodeLlama} series. Additionally, \textit{GPT-4o-mini} lacks sufficient understanding of \textit{free} command, leading to misjudgments regarding the position of \textit{shared}. Ultimately, none of LLMs are able to pass this subject.

\section{Related Work}

To assess the code generation capabilities of models, numerous code benchmarks were introduced such as APPS \citep{hendrycksMeasuringCodingChallenge2021} and CodeContests \citep{li2022competition}, which are datasets sourced from algorithmic design competitions that emphasize logical thinking. MBPP \citep{austinProgramSynthesisLarge2021} and HumanEval \citep{chenEvaluatingLargeLanguage2021} are datasets crafted manually for testing. These datasets are designed for evaluation, featuring highly independent functions and simplistic, idealized problem scenarios.
Standalone functions are predominantly focused on by these benchmarks; however, non-standalone functions are commonly encountered in pragmatic code generation scenarios \citep{yuCoderEvalBenchmarkPragmatic2024}.

Some other benchmarks attached importance on real-world problems by sourcing data from real scenarios such as StackOverflow, GitHub, and curating data manually to form benchmark datasets. For instance, PandasEval, NumpyEval \citep{zan2022cert}, and SecurityEval \citep{siddiq2022securityeval} are tailored benchmarks for specific scenarios \cite{zan2023large}. Focusing on specific scenarios is their limitation.

CoderEval \citep{yuCoderEvalBenchmarkPragmatic2024}, ClassEval \citep{du2023classeval}, DevEval \citep{li2024deveval}, and ODEX \citep{wang2023executionbased} were constructed in open domains, but they required significant investment of human labor in the stages of data curating, filtering, or annotation.

RepoEval \citep{zhangRepoCoderRepositoryLevelCode2023}, MultiPL-E \citep{multiPL}, and Exec-CSN \citep{xie2024codebenchgen} were constructed with little human involvement. However, RepoEval is focused on repository level task, which differs from ours. 
MultiPL-E obtained its data by translating other code datasets, and thus, it was not oriented towards open domains.
Exec-CSN employed LLMs to curate the CodeSearchNet dataset and generate test cases, resulting in a final dataset that has a gap from the real-world code on GitHub.

Moreover, previous benchmarks have not explored the difference of code generation capability of LLMs in multiple domains. Though a recent work~\cite{zhuoBigCodeBenchBenchmarkingCode2024} exercised capability of LLMs in using APIs from several domains, but its benchmark need human-LLM collaboration to construct. Our \name benchmark is designed not for tool utilization, but for concrete implementation of APIs and functions. Furthermore, \name can be constructed through a fully automated pipeline across open domains.
\section{Conclusion}
We introduce \textbf{\name}, a function code generation benchmark across multiple programming domains. Our research highlights the importance of developing a comprehensive benchmark to compare and assess the code generation abilities of LLMs, both in general and in specific vertical domains. The automated pipeline we introduce not only ensures diversity and real-time updates in the benchmark dataset but also facilitates the construction of custom domain benchmarks for other entities. Our preliminary findings indicate that while LLMs demonstrate remarkable performance in computation domain, their abilities in cryptography and system domains need improvement.

For the future, several research directions merit exploration. First, developing specialized training strategies and data augmentation techniques to address the shortcomings of LLMs in generating code for specific domains, such as cryptography and system. Second, utilizing our automated pipeline to construct benchmarks for a broader range of private and domain-specific code, aiming to assess and enhance LLMs' code generation performance in these areas.

\bibliography{aaai25}
\appendix
\newpage
\section{Appendix}
\subsection{List of Banned Keywords for Filtering}
\label{section:banned_keywords}

\begin{table}[ht]
    \centering
    \small
    \begin{tabular}{lll}
    \hline
    \textbf{No.} & \textbf{Banned Keyword} & \textbf{Description} \\
    \hline
    1 & os.kill & Terminate process \\
    2 & terminate & Terminate execution \\
    3 & subprocess.call(['kill', & \makecell[l]{Call subprocess to\\terminate command} \\
    4 & subprocess.call(['rm', & \makecell[l]{Call subprocess to\\delete file command} \\
    5 & subprocess.call(['rmdir', & \makecell[l]{Call subprocess to\\delete directory command} \\
    6 & subprocess.call(["kill", & \makecell[l]{Call subprocess to\\terminate command} \\
    7 & subprocess.call(["rm", & \makecell[l]{Call subprocess to\\delete file command} \\
    8 & subprocess.call(["rmdir", & \makecell[l]{Call subprocess to\\delete directory command} \\
    9 & sys.exit & System exit \\
    10 & os.unlink & Delete file \\
    11 & .unlink & Delete file \\
    12 & .rmdir & Delete directory \\
    13 & os.remove & Delete file \\
    14 & os.removedirs & Recursively delete directories \\
    15 & os.rmdir & Delete directory \\
    16 & os.system & Execute system command \\
    17 & rmtree & Recursively delete directory \\
    18 & send2trash & Send to trash \\
    19 & open( & Open file \\
    20 & .read( & Read file content \\
    21 & .write( & Write file content \\
    22 & .load( & Load data \\
    23 & .dump( & Save data \\
    24 & shutil. & File operation module \\
    25 & glob. & File matching module \\
    26 & os.path. & Path operation module \\
    27 & os.remove( & Delete file \\
    28 & os.rename( & Rename file \\
    29 & os.rmdir( & Delete directory \\
    30 & os.mkdir( & Create directory \\
    31 & os.makedirs( & Recursively create directories \\
    32 & os.listdir( & List directory contents \\
    33 & .readlines( & Read multiple lines content \\
    34 & .writelines( & Write multiple lines content \\
    35 & .seek( & File pointer positioning \\
    36 & .tell( & File pointer position \\
    \hline
    \end{tabular}
    \label{tab:keywords}
    \caption{List of Banned Keywords for Filtering}
\end{table}

In the subsection \textbf{\textit{Test-Method Matching \& Selection}}, we maintain a list of banned keywords to mitigate potential attacks that the evaluation system may suffer during code execution. If any of banned keyword appears in the code, we choose to discard this code data to prevent potential risks.

\end{document}